\documentclass[11pt,a4paper]{article}
\usepackage[hyperref]{emnlp2020}
\usepackage{times}
\usepackage{latexsym}

\usepackage{microtype} 
\usepackage{amsmath}
\usepackage{booktabs}
\usepackage{graphicx}
\usepackage{multirow}

\aclfinalcopy 

\title{Extracting Chemical--Protein Interactions via Calibrated Deep Neural Network and Self-training}

\author{Dongha Choi and Hyunju Lee \\
  Data Mining and Computational Biology Lab. \\
  School of Electrical Engineering and Comuter Science and Artificial Intelligence Graduate School\\
  Gwangju Institute of Science and Technology, Gwangju, Republic of Korea \\
  \texttt{\{dongha528,hyunjulee\}@gist.ac.kr} \\ }

\date{}

\begin{document}
\maketitle
\begin{abstract}
The extraction of interactions between chemicals and proteins from several biomedical articles is important in many fields of biomedical research such as drug development and prediction of drug side effects. Several natural language processing methods, including deep neural network (DNN) models, have been applied to address this problem. However, these methods were trained with hard-labeled data, which tend to become over-confident, leading to degradation of the model reliability. To estimate the data uncertainty and improve the reliability, ``calibration'' techniques have been applied to deep learning models.
In this study, to extract chemical--protein interactions, we propose a DNN-based approach incorporating uncertainty information and calibration techniques. Our model first encodes the input sequence using a pre-trained language-understanding model, following which it is trained using two calibration methods: mixup training and addition of a confidence penalty loss. Finally, the model is re-trained with augmented data that are extracted using the estimated uncertainties. Our approach has achieved state-of-the-art performance with regard to the Biocreative VI ChemProt task, while preserving higher calibration abilities than those of previous approaches. Furthermore, our approach also presents the possibilities of using uncertainty estimation for performance improvement.
\end{abstract}

\section{Introduction}
In the biomedical domain, there exist several entities, such as genes, chemicals, and diseases, that are closely related to each other. Therefore, extracting the relationships among these entities is critical for biomedical research, particularly in fields such as construction of a knowledge base or drug development. Biomedical text data, including PubMed abstracts, usually contain information about biomedical entities and their relationships with each other. Thus, various natural language processing models, particularly deep learning models, are applied to biomedical text data to extract the relationships among these entities, as a kind of classiﬁcation task. 

ChemProt corpus \citep{krallinger2017overview} is the first corpus dataset for chemical--protein (gene) relationship extraction, which has been conducted by BioCreative VI organizers. These organizers annotated all entity offsets of chemical and protein mentions and relationship types between chemicals and proteins (Chemical-Protein Relations, CPR). There exist 10 groups of the relationship types, and five of these (CPR:3, CPR:4, CPR:5, CPR:6, and CPR:9) were used in the evaluation. 

All models for extracting relationships from ChemProt data are designed as classifiers. In a deep learning-based multi-class classifier, the output probability distribution for each class is calculated through the Softmax function. In the training step, the model is trained to maximize the output probability of the correct class. However, some studies reported that the deep learning classifier trained with hard-labeled data (1 for correct class, 0 for else) tends to become over-confident \citep{nixon2019measuring, thulasidasan2019mixup}. This over-confidence does not directly affect classification performance, but it degrades the reliability of the model. In other words, the output probability of the over-confident model does not indicate how uncertain the input example is, even if its classiﬁcation performance is high. Therefore, several approaches, called ``calibration'' techniques, have been applied to several domains that require high reliability, such as autonomous driving and medical diagnosis  \citep{guo2017calibration, jiang2012calibrating}. 

In the natural language processing domain, bidirectional encoder representation from transformers (BERT) \citep{devlin2018bert} was proposed for a wide-range of language understanding. BERT is a large multi-head attention \citep{vaswani2017attention} model, which was pre-trained with a vast amount of corpus data. This pre-trained model can be easily transfer-learned and can be applied on several downstream tasks (e.g. sentence classification) by fine-tuning it. BERT has been used in many domains, including a biomedical field. Nevertheless, it is still important to improve the performance of BERT by applying additional techniques while using the BERT as a backbone architecture.

In this study, we propose a DNN-based approach to improve the performance of chemical--protein relationship extraction, while calibrating the classifier. More precisely, we incorporated two main calibration techniques to BERT \citep{devlin2018bert} to improve the reliability and performance. Furthermore, we propose a semi-supervised learning workflow using the calibrated model and unlabeled in-domain data. The main contributions of our study are as follows:

\begin{enumerate}
  \item We show that the additional pre-training steps of BERT with in-domain unlabeled sentences can improve the performance in a single-sentence classification task. This approach is highly applicable in the biomedical domain, since a large amount of unlabeled data is collected from PubMed and PubMed Central (PMC) using named-entity recognition models. 
  \item To the best of our knowledge, this is the first study applying calibration techniques for relationship extraction tasks. Furthermore, we also propose a training framework to apply the uncertainty information from the calibrated model to improve the performance in classification tasks. 
\end{enumerate}

\section{Related Works}
\subsection{Chemical--Protein Relationship Extraction (ChemProt)}
The relationship extraction task is a kind of text classification task, as the model should decide the relationship type of a given sentence by extracting certain semantics. Thus, several approaches including deep learning-based models are applied on the ChemProt corpus dataset. 

Some studies have reported that the syntactic dependency graph of a text sentence contains condensed and crucial information for relationship extraction, as it can be derived in the form of the shortest dependency path between two target entities \citep{bunescu2005shortest}. For example, \citet{sun2019chemical} and \citet{antunes2019extraction} proposed a chemical--protein relationship extraction model using the shortest dependency path information. \citet{wang2020graph} applied a graph convolutional network, which can capture contextual and syntactic information from text by applying a graph convolution operation on the dependency graph of the given text. 

Word representation plays a crucial role in the low level of natural language-understanding models, since it is directly related to the actual meaning of the word. Recently, \citet{peters2018deep} proposed contextualized word embedding, which can capture the meaning of words in context, as opposed to the former static word embeddings. \citet{zhang2019chemical} and \citet{sun2019deep} applied the contextual word embedding in the chemical--protein relationship extraction model, which was based on a bidirectional long short-term memory model and the attention mechanism. 

\subsection{BERT}
BERT \citep{devlin2018bert} is a large Transformer \citep{vaswani2017attention} based language understanding model, pre-trained with a vast amount of corpus data. Each layer of BERT contains a multi-head attention module and a feed-forward module. In the pre-training step, the masked language model and next-sentence prediction are applied as unsupervised learning methods. In practice, the BERT model can be fine-tuned with small in-domain data for various downstream tasks, such as text classification or question answering. The fine-tuned BERT showed state-of-the-art performances in several downstream tasks, such as natural language inference or sentiment analysis. 

\citet{lee2020biobert} generated BioBERT by pre-training the BERT model with biomedical domain corpora, namely, PubMed abstracts and PMC full-text articles. After the pre-training, the authors conducted fine-tuning regarding several biomedical domain downstream tasks, such as named entity recognition or relationship extraction. As a result, they achieved state-of-the-art performances in some tasks, which could not be achieved with the normal BERT. 

BERT and BioBERT have been applied by several researchers for the chemical--protein relationship extraction task. \citet{peng2020empirical} applied multi-task learning to BERT for biomedical domain tasks. \citet{sun2020attention} proposed a BERT-based capsule network, which uses BERT to extract long-range contexts and feed them into the attention-capsule module, and they achieved significant improvement compared to BioBERT

\subsection{Uncertainty Estimation and Calibration in Deep Learning}
To estimate the performance of the deep learning-based classifier, accuracy and F1-score are commonly used. However, these metrics only consider the correctly predicted class, regardless of the actual probability value. The expected calibration error (ECE) \citep{naeini2015obtaining} is proposed to address this issue. To calculate the ECE, all predictions are partitioned into a fixed-size bin, and the difference between the accuracy and confidence for each bin is calculated and weight-averaged. As described by \citet{guo2017calibration}, the formula for calculating the ECE and confidence is as follows: 

\begin{displaymath}
    \textrm{acc}(B_m)=\frac{1}{{|B_m|}}\sum_{i\in{B_m}}\mathbf{1}(\hat{y_i}=y_i),
\end{displaymath}
\begin{displaymath}
    \textrm{conf}(B_m)=\frac{1}{{|B_m|}} \sum_{i\in{B_m}}\hat{p_i},
\end{displaymath}
\begin{displaymath}
    \textrm{ECE}=\sum_{m=1}^{M}\frac{|B_m|}{n} |\textrm{acc}(B_m)-\textrm{conf}(B_m)|,
\end{displaymath}
where $B_m$ is the $m$-th bin, $\hat{y_i}$ is the predicted label of the $i$-th sample in the bin, and $\hat{p_i}$ is the prediction probability. $n$ is the number of test examples. The ECE of an ideally calibrated model is zero, since its accuracy and confidence are the same for every bin. 

To calibrate the deep learning classifier, several approaches have been applied. \citet{pereyra2017regularizing} approached this issue as a regularizing problem and directly applied an additional penalizing term into the loss function. Because the higher entropy of output distribution indicates over-confidence, they added negative entropy loss to a standard negative log-likelihood loss function. \citet{guo2017calibration} applied several existing post-processing calibration methods, including histogram binning and temperature scaling. To deal with the over-confidence problem caused by hard labels, label-smoothing techniques have also been applied. Label smoothing was used to improve the performance of the model as a regularization method from \citet{szegedy2016rethinking}. However, some researchers including \citet{muller2019does} proved that label smoothing can also act as a model calibration technique. 

\section{Methodology}
\subsection{Pre-processing}

\begin{figure}[h]
    \centering
    \includegraphics[scale=0.3]{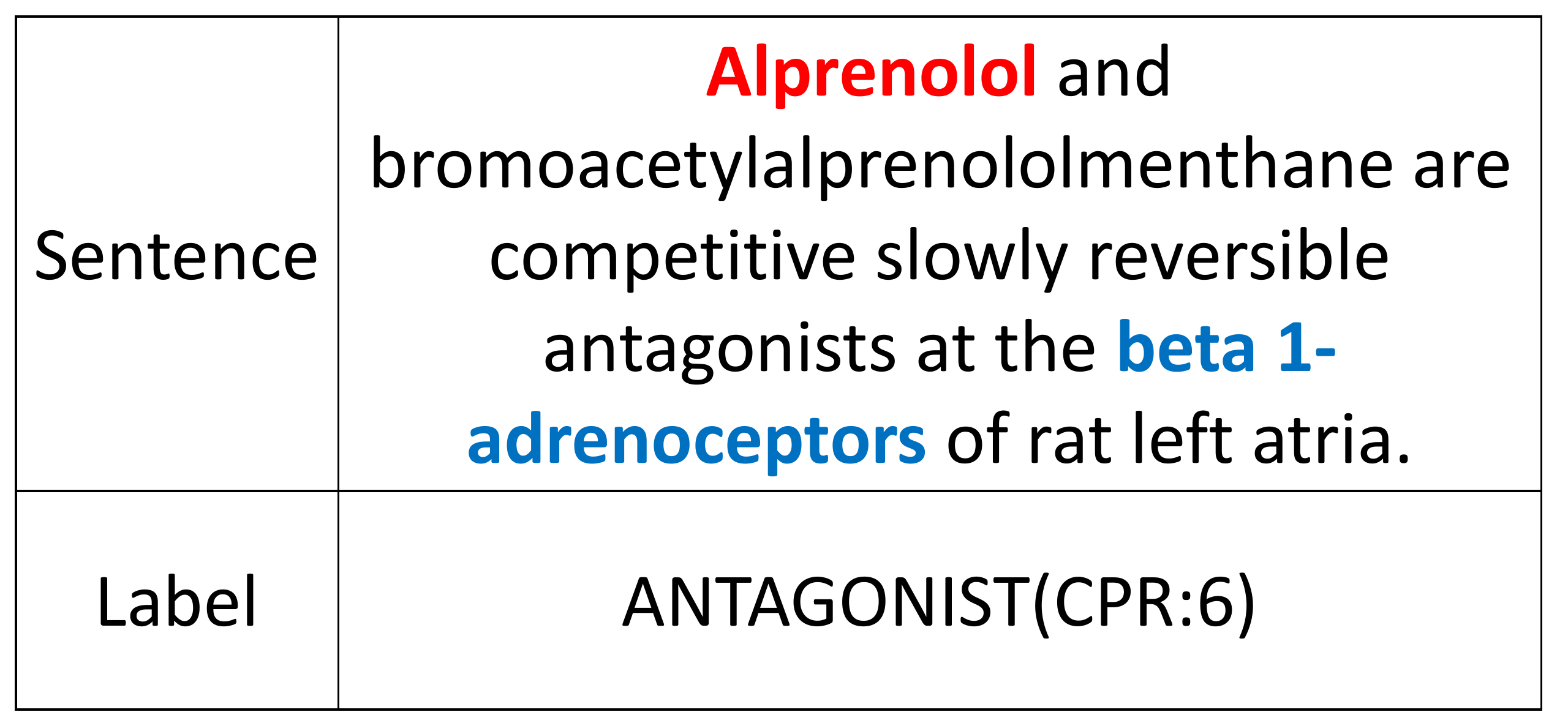}
    \caption{An example of the ChemProt data. Both entities, chemical and proteins, are highlighted in red and blue, respectively.}
    \label{fig:Figure}
\end{figure}
Each example of the ChemProt corpus dataset contains a PubMed abstract, manually annotated chemical and protein entity mentions, and gold-standard relationship labels for some protein--chemical pairs. For simplification, this dataset was pre-processed into a single-sentence task. On the basis of the entity offsets, every protein--chemical co-occurred sentences was extracted from the abstract and every sentence was labeled  with the gold-standard label or the `false' label. When there were some cross-sentence entity pairs, these were omitted for simplicity. 

It should be noted that the entity names are too speciﬁc in relationship extraction tasks and occur only a few times in the overall dataset. This means that the model vocabulary cannot contain every entity name, and their occurrence can distort the model classiﬁcation performance. To avoid this, some researchers reported that anonymizing those entity names with a pre-defined token improved the performance of the relationship extraction model \citep{lee2020biobert}. Since the ChemProt dataset provides the offset of each entity, we replaced every entity with special tokens (@GENE\$, @CHEMICAL\$). The statistics of the pre-processed ChemProt dataset are shown in Table 1. Relations belonging to each class are described in \citet{krallinger2017overview}, and representative relations of CPR:3, CPR:4, CPR:5, CPR:6, and CPR:9 are upregulator, downregulator, agonist, antagonist, and substrate, respectively. Figure 1 shows an example of the ChemProt data. The original version of ChemProt data can be downloaded from https://biocreative.bioinformatics.udel.edu/.

\begin{table}[h]
\centering
\renewcommand{\tabcolsep}{0.8mm}
\begin{tabular}{lllllll}
\toprule
\small{Dataset}     & \small{CPR:3} & \small{CPR:4} & \small{CPR:5} & \small{CPR:6} & \small{CPR:9} & \small{False} \\
\midrule
Train       & 757   & 2233  & 170   & 229   & 727   & 13749 \\
Dev & 546   & 1092  & 115   & 199   & 457   & 8854  \\
Test        & 662   & 1637  & 182   & 293   & 642   & 12167 \\
Total       & 1965  & 4962  & 467   & 721   & 1826  & 34770 \\
\bottomrule
\end{tabular}
\caption{Statistics of pre-processed ChemProt dataset with evaluated labels}\label{tab:accents}
\end{table}

\subsection{Additional Pre-training of BERT with In-domain Sentences}
As mentioned earlier, BERT and BioBERT could understand the deep contextual information of the text through pre-training with huge corpora. In particular, BioBERT suggested that the pre-training of BERT in a specific domain may contribute to performance improvement in a downstream task of the domain. Furthermore, some research has shown that additional pre-training of BERT with an unlabeled corpus dataset from the same domain can improve the performance of specific downstream tasks \citep{xie2019unsupervised, sun2019fine}. Similarly, to pre-train BioBERT for chemical---protein relationship extraction, we collected protein-chemical co-occurred sentences from PubMed abstracts using PubTator \citep{wei2013pubtator}. As a result, a total of 8.2 million protein--chemical co-occurred sentences were collected. Subsequently, we pre-trained BioBERT with collected sentences. Unlike the pre-training process of the original BERT and BioBERT, this process involved each input being fed as a single sentence. Although this approach is not suitable for next-sentence prediction in BERT, we did not consider it as some studies have shown the ineffectiveness of next-sentence prediction \citep{liu2019roberta, yang2019xlnet}. This additional pre-training is so domain-specific that it may distort the original ability to capture the deep contextual information. Thus, we employed a learning rate smaller than that of the original process employed in BERT and BioBERT. The pre-training is performed over a total of 1 million steps with a batch size of 40. 

\subsection{Calibration Methods}
\subsubsection{Mixup training}
Mixup training is a data-augmentation method originally applied in the computer vision domain \citep{zhang2017mixup}. In this method, two random examples and their labels are convexly combined in a random ratio. More precisely, mixup training can be shown as per the formula below:

$$\begin{array}{c}
    \tilde{x}=\lambda{x_i}+(1-\lambda){x_j},\\
    \tilde{y}=\lambda{y_i}+(1-\lambda){y_j},
\end{array}$$
where $x_i$ and $x_j$ are two randomly sampled examples and  $y_i$ and $y_j$ are their one-hot labels. $\lambda$ is the randomly generated mixing ratio. In this study, every random sampling in the mixup training was performed under uniform distribution. 

\begin{figure}[t]
    \centering
    \includegraphics[scale=0.5]{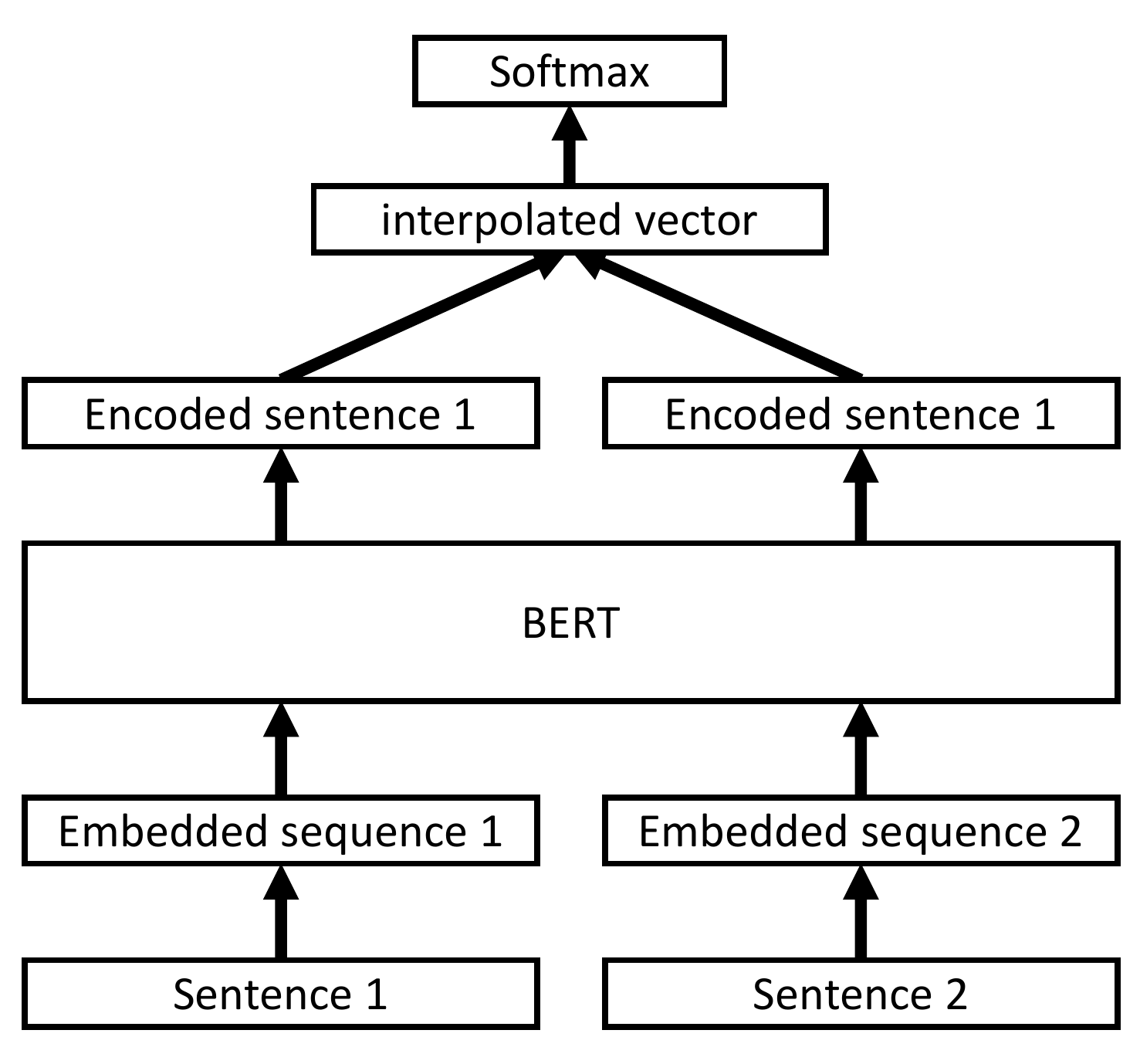}
    \caption{Architecture of the mixup BERT model}
    \label{fig:Figure}
\end{figure}
Although mixup training is proposed for regularization and performance-boosting in the computer vision domain, \citet{thulasidasan2019mixup} reported that training a deep learning model using the mixup method shows more calibrated results. They applied mixup training to well-known DNN models to show their calibration ability in several domains, including image and sentiment classification. Particularly in text classification, as input-level mixing is not possible regarding text input, mixup architecture was used for text input, as proposed by \citet{guo2019augmenting}. In order to mix the input sequence of the token itself, the embedded sequence of the input or the encoded feature vector from the neural network was mixed. Similarly, in this study, we applied architecture from \citet{guo2019augmenting}, with BERT as the sentence encoder. We used the classification embedding (`[CLS]’) vector of BERT as an encoded sentence, as it can be processed as a compressed representation of the overall sentence in classification tasks. As illustrated in Figure 2, two input sequences are fed into the BERT model, and two feature vectors encoded by BERT are convexly interpolated as per a given mixing ratio. Then, the mixed vector is fed into the Softmax layer to generate output probability distribution. In this step, the output of the model should predict the ratio between two classes instead of a single correct label. In specific, our mixup model architecture can be formulated as:
$$\begin{array}{c}
    \tilde{x}=\lambda{f(x_i)_{[CLS]}}+(1-\lambda){f(x_j)_{[CLS]}} \\
    \tilde{y}=\lambda{y_i}+(1-\lambda){y_j},
\end{array}$$
where $f_{[CLS]}$ denotes the classification embedding of the last layer activation in BERT.

\subsubsection{Confidence Penalty Loss (CPL)}
We applied an additional calibration technique not only in the model architecture but also on the loss function as a regularization term. As \citet{pereyra2017regularizing} proposed, we applied the penalizing term of low entropy output distribution to our model. As mentioned earlier, the output distribution of the uncalibrated model is biased towards 0 and 1. In other words, the output of the uncalibrated model has a low entropy value. Thus, the incorporation of negative entropy to the original loss function can enable the functioning as a regularization term for calibration in the training step. More precisely,  when the output probability distribution is written as $p_{\theta}(y|x)$, the entropy of the output probability distribution can be expressed as:
\begin{displaymath}
    H(p_{\theta}(\tilde{y}|\tilde{x}))=-\sum_{i}p_{\theta}(\tilde{y_i}|\tilde{x})\log(p_{\theta}(\tilde{y_i}|\tilde{x})),
\end{displaymath}
where $i$ indicates the index of each class. The final classification loss of our model is defined as the weighted sum of the standard classification loss and negative entropy, 
\begin{displaymath}
    J(\theta)=-\sum{p_{\theta}(\tilde{y}|\tilde{x})}-{\beta}H({p_{\theta}(\tilde{y}|\tilde{x})})
\end{displaymath}
with the hyper-parameter $\beta$, which controls the strength of the penalty for over-confidence. 

Even though the original BERT applied dropout on the classification layer, we excluded dropout because the overlap of the regularization method may cause underfitting. By incorporating a confidence penalty loss term, we expect the model to avoid over-confidence and achieve even better generalization as the effect of regularization.

\subsection{Self-training}
\begin{figure}[t]
    \centering
    \includegraphics[scale=0.64]{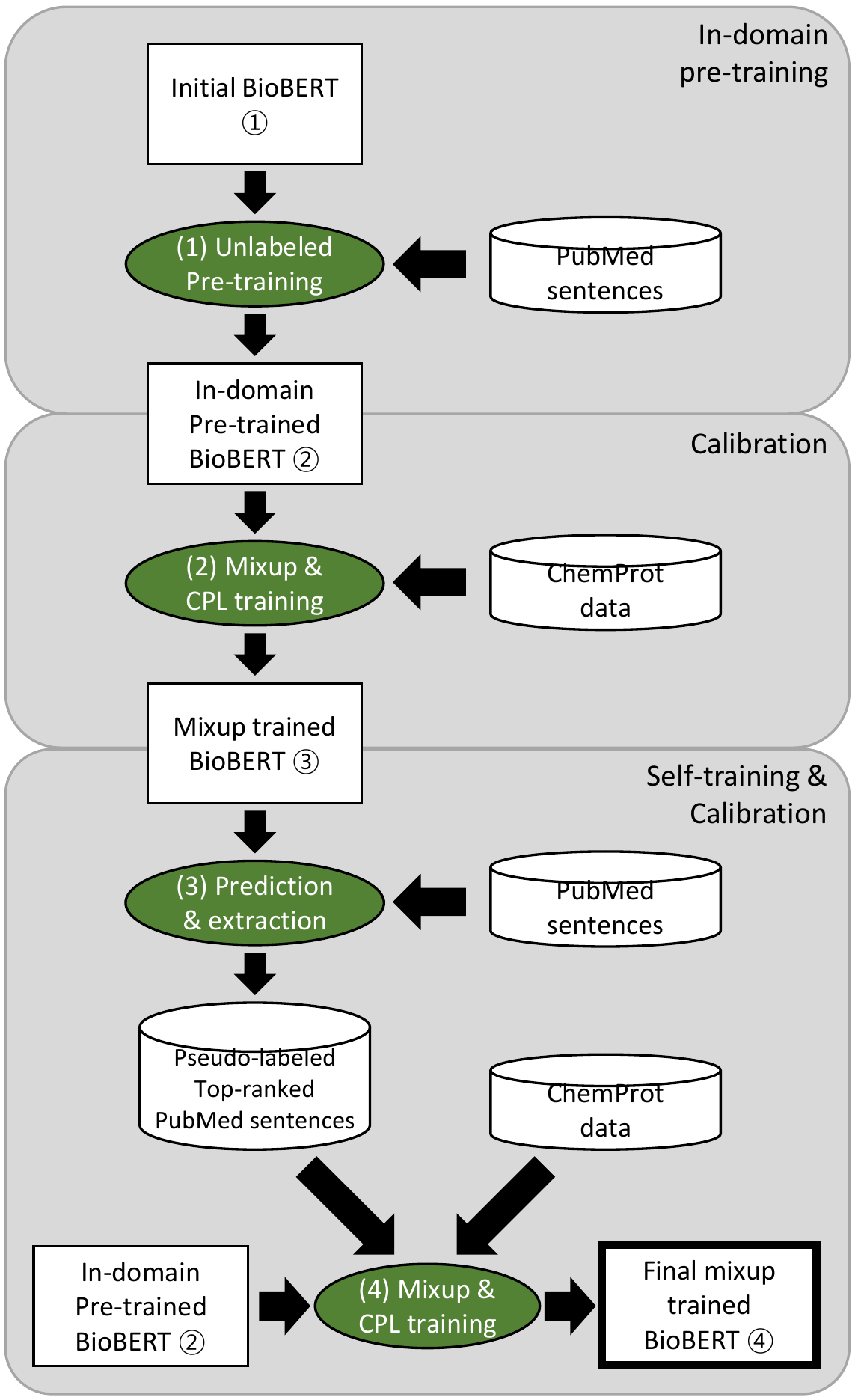}
    \caption{Overall workflow of self-training with the calibrated model. 1) An initial BioBERT is pre-trained with protein-chemical co-occurred unlabeled sentences. 2) A pre-trained BioBERT is fine-tuned by mixup training and CPL using ChemProt data. 3) The output probabilities of PubMed unlabeled sentences are predicted with the third model, and top-$k$ sentences are extracted and then pseudo-labeled. 4) The BioBERT model is fine-tuned by mixup training and augmented data, generating the final model. Circled numbers are used to distinguish the different models. }
    \label{fig:Figure}
\end{figure}
Self-training is a kind of semi-supervised learning method, which allows for the labeled dataset to be enlarged using unlabeled data within a similar domain \citep{triguero2015self}. The basic principle of self-training is to label some of the unlabeled data through a model trained with labeled data. When a certain level of performance is guaranteed, unlabeled data predicted with a high probability in the classifier is likely to have a corresponding label. Thus, such data can be used as pseudo-labeled data, even if it contains slight noise. Furthermore, the calibrated model may show a more reliable prediction probability than that of the over-confident model and even derive more qualified pseudo-labeled data. 

\begin{table*}[h]
\centering
\begin{tabular}{lllll}
\toprule
Models        & \# of augmented data & F1-score(\%) & ECE    & OE     \\ 
\midrule
BioBERT only & 0                                                                  & 77.15        & 0.0841 & 0.0827 \\
Our model (k=0)          & 0                                                                  & 78.34        & 0.0298 & \textbf{0}      \\
Our model (k=200)        & 9000                                                               & \textbf{78.83}        & 0.0232 & 0.0005 \\
Our model (k=400)        & 18000                                                              & 78.42        & 0.0128 & 0.0021 \\
Our model (k=600)        & 27000                                                              & 78.30        & \textbf{0.0102} & 0.0051 \\ 
\bottomrule
\end{tabular}
\caption{Calibration performance of our proposed models. }\label{tab:accents}
\end{table*}

\begin{figure*}[h]
    \centering
    \includegraphics[scale=0.35]{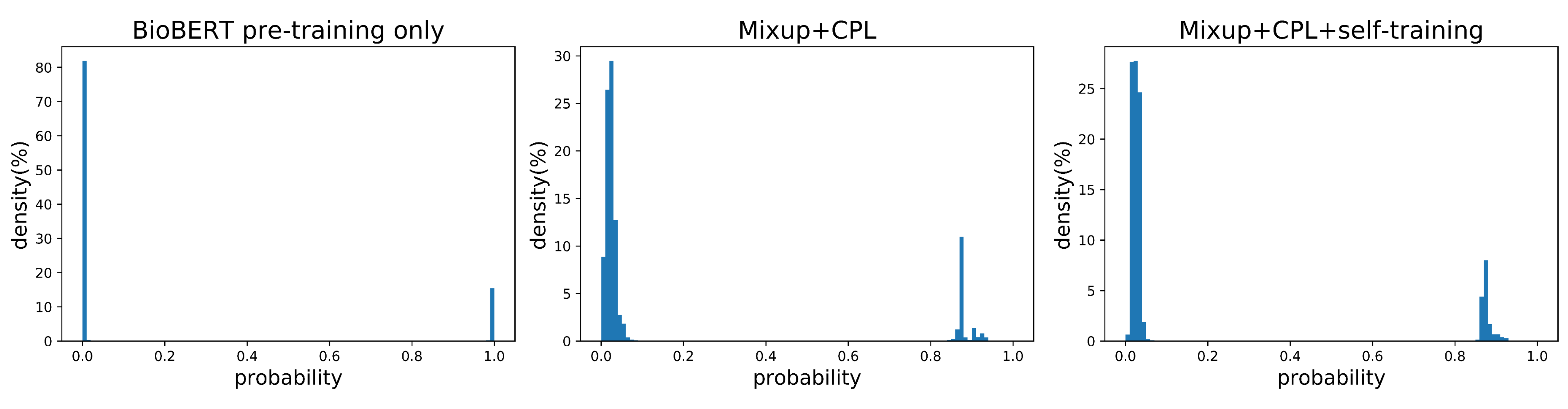}
    \caption{Histogram of output probabilities. \textbf{Left:} Normal BioBERT model pre-trained with in-domain sentence. \textbf{Middle:} Mixup trained model with confidence penalty loss. \textbf{Right:} Self-trained final model.}
    \label{fig:Figure}
\end{figure*}

\begin{table*}[!h]
\centering
\begin{tabular}{llllll}
\toprule
Models                                                                              & F1-score(\%) & Accuracy(\%) & Confidence(\%) & ECE    & OE     \\ 
\midrule
BioBERT  (our experiment)                                                       & 77.15        & 89.92        & 98.31          & 0.0841 & 0.0827 \\
BioBERT+PT                                                                         & 77.68        & 90.40        & 98.44          & 0.0806 & 0.0793 \\
BioBERT+PT+mixup                                                                   & 77.92        & 90.18        & 97.70          & 0.0754 & 0.0737 \\
BioBERT+PT+CPL                                                                     & 78.18        & 90.57        & 88.14          & 0.0241 & \textbf{0}      \\
BioBERT+PT+mixup+CPL                                                               & 78.34        & 90.54        & 87.53          & 0.0298 & \textbf{0}      \\
\begin{tabular}[c]{@{}l@{}}BioBERT+PT+mixup+CPL+ST\\ (Proposed model)\end{tabular} & \textbf{78.83}        & 90.13        & 87.90          & \textbf{0.0232} & 0.0005 \\ 
\bottomrule
\end{tabular}
\caption{Results of ablation study. (PT: Pre-training, CPL: Confidence penalty loss, ST: Self-training)}\label{tab:accents}
\end{table*}

\begin{table}[h]
\centering
\begin{tabular}{llll}
\toprule
Models                                                       & P(\%) & R(\%) & F(\%) \\
\midrule
BERT                                                         & 74.01 & 70.79 & 72.36 \\
BioBERT (paper)     & 76.63 & 76.74 & 76.68 \\
BioBERT (our exp.) & 77.64 & 76.82 & 77.15 \\
\citet{lim2018chemical}                                                            & 74.80 & 56.00 & 64.10 \\
\citet{sun2019chemical}                                                            & 77.08 & 76.06 & 76.56 \\
\midrule
\midrule
Our best model                                               & \textbf{77.76} & \textbf{80.10} & \textbf{78.83} \\
\bottomrule
\end{tabular}
\caption{Comparison between our proposed models and other models on the basis of classification performance. }\label{tab:accents}
\end{table}

In this study, we used 8.2 million gene--chemical co-occurred sentences from PubMed, extracted previously as unlabeled data. We first fine-tuned our model with the labeled data (ChemProt) and then performed the prediction of unlabeled data using the fine-tuned model. After the prediction, we extracted the examples with top-$k$ probability per 1 million examples for every label except for `false'. We excluded this label because the label distribution of ChemProt dataset is biased to `false'. Finally, we fine-tuned a new model with original and pseudo-labeled data. Figure 3 shows the overall workflow of the self-training process involving the calibrated model. 

\section{Results}

\subsection{Experimental Setup and Hyperparameters}
We use a BERT-base-cased network with a BioBERT weight pre-trained using PubMed and PMC as the sentence encoder. Same as the original BERT-base model, our model contains 110 million parameters. Each pre-training and fine-tuning experiment is performed using an NVIDIA Titan Xp 12GB GPU. It takes about four hours to fine-tune our model with only ChemProt data, and it takes 1.5 hours more for the increase of 9,000 sentences for self-training. The length of the sentence fed into the sentence encoder model is set to 200, since $<$ 1\% of ChemProt sentences exceed the length of 200. During mixup training, we generated three additional examples per sentence via random sampling of the sentence and mixup ratio with uniform distribution. We set the weight of the confidence penalty loss ($\beta$) to 0.3 because the model showed the best calibration in the development set with $\beta=0.3$ in the grid search on [0.1, 0.3, 0.5]. The hyper-parameter tuning results in the development set are shown in the Appendix section. 

\subsection{Calibration evaluation}
To evaluate the calibration performance of our approach, we compared each of our techniques on the basis of the ECE and over-confidence error (OE), which is defined by \citet{thulasidasan2019mixup}. The OE is formulated as:

{\scriptsize
\begin{displaymath}
\begin{split}
    \textrm{OE}=\sum_{m=1}^{M}{\frac{|B_m|}{n}}[\textrm{conf}(B_m)\times\textrm{max}(\textrm{conf}(B_m)-\textrm{acc}(B_m),0)]
\end{split}
\end{displaymath}}

As shown in Table 2, the models trained using mixup training and confidence penalty are calibrated better than normal BioBERT models. Notably, self-training also improved the calibration ability, as reflected by the ECE values. \citet{thulasidasan2019mixup} reported that mixing the labels during mixup training can function as label smoothing, and they emphasized that label smoothing plays a crucial role in terms of calibration. 
The augmentation of additional labeled sentences via self-training also implies that label-smoothened mixup examples are augmented. Thus, a self-trained model with a higher $k$ showed better calibration results. However, performances of models with $k >200$ were degraded, and the performance of the model with $k$ = 600 performed worse than that of the model with $k$ = 0 . This might be because the augmentation of too much data may include irrelevant sentences into the training set, which can distort the characteristics of the original data. This means that the determination of $k$ during self-training presents a trade-off between classification performance and calibration performance. Additionally, Figure 4 visualizes the qualitative calibration effect of our approach.

\subsection{Ablation study}
To observe how each technique contributed to the overall performance improvement, we conducted an ablation study. We ablated in-domain pre-training, mixup training, confidence penalty loss, and self-training progressively, the results of which are shown in Table 3. The in-domain pre-training improved classification performance compared to BioBERT, but it did not show improvement in terms of calibration. Similar to the results shown by \citet{guo2019augmenting} and \citet{thulasidasan2019mixup}, mixup training improved both the F1-score and calibration score. Regularization with confidence penalty loss also yielded significant improvement in terms of both classification and calibration, and the simultaneous application of mixup and CPL also showed positive results. As mentioned earlier, the application of self-training on the calibrated model enhanced performance overall, except for the aspect concerning over-confidence of error. 

\subsection{Performance Comparison}
We compared the classification performance of our model with that of several chemical--protein relationship extraction models. As shown in Table 4,  BioBERT experimented by us showed a slightly higher F1 score than that of BioBERT. As the other hyper-parameters were the same as those of the original conditions, it appears that there was a slight improvement in performance owing to the increase in the input sequence length (128 in the case of BioBERT).  Our models outperformed BioBERT---the current state-of-the-art model in ChemProt. As shown in Table 3, the mixup+CPL model, trained with only the original ChemProt dataset, also performed better than BioBERT. 

\section{Conclusion}
In this study, we propose a calibrated deep neural network-based relationship extraction model for chemical--protein interactions. We applied mixup training---which can both augment the training examples and calibrate the model---on the BioBERT model. We also incorporated the low entropy penalizing term in the loss function, as a regularization term during training. This led to significant improvement in terms of both classification and calibration performance. Moreover, we applied self-training on our model to augment the training data and boost the performance, as our model is well-calibrated; it returned reliable output probabilities. Consequently, our model outperformed the other chemical--protein relationship extraction models and achieved state-of-the-art performance regarding the Biocreative VI ChemProt task. 

In this work, we applied our training process on the biomedical domain, especially on chemical--protein relation extraction, because a large set of unlabeled data can be found from PubMed in the biomedical domain. This process can be applied to any NLP classification problems with unlabeled data sets. We can apply the proposed method to other tasks such as  CoLA or SST-2 in the GLUE benchmark \citep{wang2018glue} since there are a large set of unlabeled data for ungrammatical sentences or movie reviews. The application of our method to other domains will be our future work.

\section*{Acknowledgments}
This research was supported by the Bio-Synergy Research Project (NRF-2016M3A9C4939665) of the Ministry of Science and ICT through the National Research Foundation of Korea (NRF) and the Institute of Information \& communications Technology Planning \& Evaluation (IITP) grant funded by the Korea government (MSIT) (No.2019-0-01842, Artificial Intelligence Graduate School Program (GIST)).

\bibliography{emnlp2020}
\bibliographystyle{acl_natbib}

\appendix

\section{Appendices}
\label{sec:appendix}
To find the best weight of the confidence penalty loss ($\beta$), we measured performance of the proposed model using a development set of ChemProt data for three values of $\beta$ = [0.1, 0.3, 0.5]. Three replicates were performed for each $\beta$, and we calculated the mean and variance of each $\beta$. The results are shown in Table A1. Since the confidence penalty loss was incorporated to the loss function for enhancing calibration, we chose $\beta=0.3$, which showed the best calibration performance. 

To show the significant improvement in terms of classification performance, we report the mean and standard deviation values of every experiment in the ablation study. Table A2 shows the statistics of all metrics in the ablation study. 

The max, min and std of F-scores of our proposed model are 79.92, 78.18 and 0.59, respectively, while those of BioBERT are 77.36, 76,91 and 0.18. The max, min and std precision of our proposed model are 80.08, 75.64 and 1.56 while those of BioBERT are 78.64, 76.18 and 0.83. The max, min and std recall of our proposed model are 81.84, 78.73 and 1.00 while those of BioBERT are 77.76, 76.13 and 0.62. Although the improvement of the precision by our model is relatively small, recall and F-score were more improved. The min values of recall and F- score of our model are larger than the max values of the BioBERT, showing superiority of our method.

\setcounter{table}{0}
\renewcommand{\thetable}{\Alph{section}\arabic{table}}

\begin{table*}[hbt!]
\centering
\begin{tabular}{lllllll}
\toprule
$\beta$ & F1-score (\%) & Variance & Accuracy (\%) & Confidence (\%) & ECE    & OE     \\
\midrule
0.1     & 81.13 & $1.340\times10^{-5}$  & 91.14 & 96.36 & 0.0522 & 0.0503 \\
0.3     & 81.19 & $1.705\times10^{-5}$  & 91.05 & 87.6  & \textbf{0.0344} & 0      \\
0.5     & \textbf{82.27} & $7.698\times10^{-7}$ & 91.37 & 78.13 & 0.1622 & 0     \\
\bottomrule
\end{tabular}
\caption{Results of hyperparameter search using the development set.}\label{table}
\end{table*}

\begin{table*}[hbt!]
\renewcommand{\tabcolsep}{0.6mm}
\centering
\begin{tabular}{ccccccccccccccc}
\toprule
\multirow{2}{*}{Model}                                             & \multicolumn{2}{c}{P (\%)} & \multicolumn{2}{c}{R (\%)} & \multicolumn{2}{c}{F (\%)} & \multicolumn{2}{c}{Acc. (\%)} & \multicolumn{2}{c}{Conf(\%)} & \multicolumn{2}{c}{ECE}  & \multicolumn{2}{c}{OE} \\
                                                                   & mean              & std.   & mean              & std.   & mean              & std.   & mean           & std.         & mean          & std.         & mean            & std.   & mean         & std.    \\ \midrule
BioBERT                                                            & 77.64             & 0.83   & 76.82             & 0.62   & 77.15             & 0.18   & 89.92          & 0.12         & 98.31         & 0.08         & 0.0841          & 0.0013 & 0.0827       & 0.0013  \\
BioBERT+PT                                                         & 79.18             & 1.08   & 76.43             & 0.71   & 77.68             & 0.39   & 90.40          & 0.22         & 98.44         & 0.11         & 0.0806          & 0.002  & 0.0793       & 0.002   \\
\begin{tabular}[c]{@{}c@{}}BioBERT+PT+\\ mixup\end{tabular}        & 78.46             & 0.65   & 77.49             & 0.69   & 77.92             & 0.53   & 90.18          & 0.18         & 97.70         & 0.07         & 0.0754          & 0.0015 & 0.0737       & 0.0015  \\
\begin{tabular}[c]{@{}c@{}}BioBERT+PT+\\ CPL\end{tabular}          & 79.96             & 0.77   & 76.61             & 0.83   & 78.18             & 0.46   & 90.57          & 0.09         & 88.14         & 0.13         & 0.0241          & 0.0015 & \textbf{0}   & 0       \\
\begin{tabular}[c]{@{}c@{}}BioBERT+PT+\\ mixup+CPL\end{tabular}    & \textbf{80.36}    & 1.19   & 76.58             & 0.57   & 78.34             & 0.52   & 90.54          & 0.20         & 87.52         & 0.12         & 0.0298          & 0.002  & \textbf{0}   & 0       \\
\begin{tabular}[c]{@{}c@{}}BioBERT+PT+\\ mixup+CPL+ST\end{tabular} & 77.76             & 1.56   & \textbf{80.10}    & 1.00   & \textbf{78.83}    & 0.59   & 90.13          & 0.39         & 87.90         & 1.16         & \textbf{0.0232} & 0.0101 & 0.0005       & 0.0011  \\ 
\bottomrule
\end{tabular}
\caption{Results of ablation study. (PT: Pre-training, CPL: Confidence penalty loss, ST: Self-training)\newline
\newline
\newline
\newline
\newline
\newline
\newline
\newline
\newline
\newline
\newline
\newline
\newline
\newline
\newline
\newline
\newline
\newline
\newline
\newline
\newline
\newline
\newline
\newline
\newline
\newline
\newline
\newline
\newline
\newline
}\label{tab:accents}

\end{table*}

\end{document}